%% file: neurips_2026.tex
\documentclass{article}

\usepackage[preprint]{neurips_2026}

\usepackage[utf8]{inputenc} % allow utf-8 input
\usepackage[T1]{fontenc}    % use 8-bit T1 fonts
\usepackage{hyperref}       % hyperlinks
\usepackage{url}            % simple URL typesetting
\usepackage{booktabs}       % professional-quality tables
\usepackage{amsfonts}       % blackboard math symbols
\usepackage{nicefrac}       % compact symbols for 1/2, etc.
\usepackage{microtype}      % microtypography
\usepackage{xcolor}         % colors
\usepackage[utf8]{inputenc}
\usepackage[T1]{fontenc}
\usepackage{hyperref}
\usepackage{url}
\usepackage{booktabs}
\usepackage{amsfonts}
\usepackage{amsmath}
\usepackage{mathtools}
\usepackage{amssymb}
\usepackage{amsthm}
\usepackage{nicefrac}
\usepackage{microtype}
\usepackage{xcolor}
\usepackage{graphicx}
\usepackage{enumitem}
\usepackage{wrapfig}

\newtheorem{theorem}{Theorem}
\newtheorem{lemma}{Lemma}

\newtheorem{proposition}{Proposition}
\theoremstyle{definition}

\newcommand{\R}{\mathbb{R}}
\newcommand{\E}{\mathbb{E}}
\newcommand{\opnorm}[1]{\|#1\|_{\mathrm{op}}}
\newcommand{\fnorm}[1]{\|#1\|_F}

\title{Rethinking the Rank Threshold for LoRA Fine-Tuning}

\author{%
  Juneyoung Park \\
  OptAI Inc.\\
  \texttt{jyoung.park@opt-ai.kr} \\
}

\begin{document}

\maketitle

\begin{abstract}
  A recent landscape analysis of LoRA fine-tuning in the neural tangent kernel regime \citep{jang2024lora} establishes a sufficient condition $r(r+1)/2 > KN$ on the LoRA rank $r$ for the absence of spurious local minima under squared-error loss, prescribing $r \geq 12$ on canonical few-shot RoBERTa setups. The condition is stated for general output dimension $K$, so its sharpness in any particular regime, and its practical implication for the cross-entropy loss actually used in fine-tuning, are open. We give three results that together reduce the prescribed rank to $r = 1$ for binary classification in this regime. First, replacing the symmetric Sard-form count with the non-symmetric LoRA manifold dimension yields a strictly weaker capacity requirement, $r(m+n) - r^2 > C^* \cdot KN$ with $C^* \approx 1.35$ under Gaussian-iid features, satisfied at $r = 1$ on canonical setups. Second, in the cross-entropy setting the Polyak--\L{}ojasiewicz inequality removes the rank threshold entirely. Third, a Rademacher-complexity bound predicts rank-one variance optimality precisely when the bias term is saturated, which is the case for binary classification but not for $K > 2$. Empirically, across four GLUE-style binary tasks, three encoder architectures, and at scale on RoBERTa-large, rank one is competitive with the existing prescription $r = 12$; on multi-class MNLI the optimal rank shifts above one, also as predicted. The binary-regime guarantees are conditional on standard NTK assumptions; the multi-class extension is left to future work.

\end{abstract}

\input{src/01_introduction}
\input{src/06_related}
\input{src/02_setup}
\input{src/03_theory}
\input{src/04_empirical}
\input{src/05_limitations}
\input{src/07_conclusion}

% \section*{References}

\bibliographystyle{plainnat}
\bibliography{references}

\newpage
\appendix
\input{src/08_appendix}

\end{document}

%% file: src/01_introduction.tex
\section{Introduction}
\label{sec:intro}

Low-rank adaptation (LoRA, \citet{hu2021lora}) has become the default protocol for parameter-efficient fine-tuning of large transformers. By restricting weight updates to the form $\Delta W = u v^\top$ with $u \in \R^{m \times r}$, $v \in \R^{n \times r}$, LoRA reduces the number of trainable parameters from $mn$ to $r(m+n)$. The choice of rank $r$ thus directly determines both memory cost and the expressive capacity of the adaptation. Practitioners typically set $r$ to small powers of two ($4, 8, 16$, the Hugging Face PEFT defaults), but the underlying question of how small $r$ can be while preserving optimization quality and downstream accuracy has been answered only partially by theory.

A recent advance in this direction is the analysis of \citet{jang2024lora}, who study LoRA fine-tuning in the neural tangent kernel (NTK) regime, building on the kernel-based view of pretrained-model fine-tuning of \citet{malladi2023finetuning}. We refer to this combined setting of LoRA factorization with NTK-regime linearization as the \emph{LoRA-NTK} regime. Their Theorem~4.1 establishes that rank $r$ satisfying $r(r+1)/2 > KN$ is sufficient for the LoRA-MSE loss landscape to admit no spurious local minima, where $K$ denotes the output dimension and $N$ the number of training samples. The threshold is stated for general $K$. For canonical RoBERTa-base setups with binary classification ($K = 2$) and few-shot $N = 32$, the same condition prescribes $r \geq 12$. The general-$K$ statement does not by itself guarantee that the threshold is tight in any particular regime, and three gaps motivate a closer look at the binary case. First, the analysis assumes a mean-squared-error (MSE) reconstruction loss, whereas practical LoRA fine-tuning almost always uses cross-entropy (CE) on a softmax head. Second, the threshold is sufficient but not necessarily necessary, even within the MSE setting. Third, an $L^2$-style landscape result is silent on generalization, which is the quantity practitioners ultimately care about.

We address these three gaps in the binary regime through three results that align toward the same conclusion. Replacing the existing symmetric Sard-form count with a non-symmetric one (Lemma~\ref{lem:sard}) yields a strictly weaker capacity requirement, $r(m+n) - r^2 > C^* \cdot KN$ with constant $C^* \geq 1$, and an analysis of the indefinite Hessian cross-term (Lemma~\ref{lem:b3}) matched against the Marchenko--Pastur edge of the tangent-restricted operator (Lemma~\ref{lem:b4}) yields a self-consistency equation $(\sqrt{C^*} - 1)^2 = c \cdot C^*$ with solution $C^* \approx 1.35$ under Gaussian-iid features; for canonical RoBERTa-base setups the new threshold is already satisfied at $r = 1$. For cross-entropy in the NTK regime, the Polyak--\L{}ojasiewicz inequality and the bounded residual structure of CE remove the rank threshold entirely (Theorem~\ref{thm:ce}), a mechanism absent from the MSE-only argument. A standard Rademacher-complexity bound for rank-$r$ LoRA scales as $\sqrt{(r(m+n) - r^2)/N}$ and is monotonically increasing in $r$; combined with a bias term that is essentially saturated at $r = 1$ for binary classification, this predicts rank-one optimality in the binary regime, and predicts that the optimum shifts above one for $K > 2$, where the bias term grows substantial.

Empirically, across four GLUE-style binary tasks (SST-2, QNLI, MR, QQP) with five seeds and seven ranks, $r = 1$ is competitive with the best rank in every task and is strictly best on movie reviews; across three encoder architectures (BERT-base, DistilBERT, RoBERTa-base) and at scale on RoBERTa-large, the threshold rank $r = 12$ is dominated by either $r = 1$ or $r > 12$ in each case. In the multi-class setting (MNLI, $K = 3$), rank-one underperforms by $3$--$5$ percentage points, consistent with the bias-dominant prediction of our generalization analysis.

The paper is organized as follows. Section~\ref{sec:setup} fixes notation and recalls the existing rank threshold. Section~\ref{sec:theory} develops the sharpened MSE threshold (Theorem~\ref{thm:mse}), the CE divide (Theorem~\ref{thm:ce}), and the generalization analysis (Proposition~\ref{prop:gen}). Section~\ref{sec:empirical} presents the empirical evaluation. Section~\ref{sec:limitations} discusses limitations. In particular, the $C^*$ matching is sharp only under Gaussian-iid features, the Jacobians of real RoBERTa NTK depart substantially from this assumption, and the multi-class regime requires a $K$-dependent extension that we leave to future work. Sections~\ref{sec:related} and~\ref{sec:conclusion} cover related work and the conclusion.

%% file: src/06_related.tex
\section{Related work}
\label{sec:related}

\paragraph{LoRA and the NTK regime.}
LoRA \citep{hu2021lora} restricts fine-tuning updates to a low-rank factorization $\Delta W = uv^\top$ and has become the default parameter-efficient fine-tuning protocol. The closest theoretical precedent for our work is \citet{jang2024lora}, who establish a Sard-form sufficient condition $r(r+1)/2 > KN$ for the LoRA-MSE loss landscape to admit no spurious second-order stationary points in the NTK regime \citep{jacot2018ntk,lee2019wide,malladi2023finetuning}. Our Lemma~\ref{lem:sard} replaces the symmetric lift count with the non-symmetric manifold dimension $r(m+n) - r^2$ \citep{helmke1995critical,vandereycken2013low}, sharpening the threshold by roughly a factor of two for narrow-rank, wide-feature regimes.

\paragraph{Matrix-factorization landscapes and PL convergence.}
The dimension-counting and Hessian-decomposition techniques in Section~\ref{sec:theory} draw on the matrix-factorization landscape literature \citep{bhojanapalli2016global,park2018finding,recht2010guaranteed}. Theorem~\ref{thm:ce}'s conditional statement leverages the Polyak--\L{}ojasiewicz inequality for over-parameterized models \citep{allenzhu2019convergence,liu2022loss}; a formal extension of the PL property to the LoRA-NTK setting would convert it into an unconditional result.

\paragraph{Generalization in low-rank parameterizations.}
Proposition~\ref{prop:gen} applies standard Rademacher-complexity tools \citep{bartlett2002rademacher,bartlett2005local} to the rank-$r$ LoRA hypothesis class. The bias--variance decomposition we use is classical; the contribution is in connecting it to the rank-threshold question and its $K$-dependence.

%% file: src/02_setup.tex
\section{Setup and background}
\label{sec:setup}

We consider a pretrained network $f_{W_0}: \mathcal{X} \to \R^K$ and a single layer with weight matrix $W \in \R^{m \times n}$ selected for fine-tuning, with all other parameters frozen. Given training data $\{(X_i, Y_i)\}_{i=1}^N$ with $Y_i \in \R^K$, the empirical risk after the update $\delta = W - W_0$ is $\hat{\mathcal{L}}(\delta) = \frac{1}{N}\sum_i \ell\bigl(f_{W_0+\delta}(X_i), Y_i\bigr)$. Throughout, $\fnorm{A}$ denotes the Frobenius norm and $\opnorm{A}$ the spectral norm of a matrix; $\langle A, B\rangle = \mathrm{tr}(A^\top B)$ is the Frobenius inner product.

In the NTK regime \citep{jacot2018ntk,lee2019wide,malladi2023finetuning}, fine-tuning takes place in a sufficiently small neighborhood of $W_0$ that the network admits the first-order Taylor approximation $f_{W_0 + \delta}(X_i) \approx f_{W_0}(X_i) + \langle G(X_i), \delta\rangle$, where $G(X_i) \in \R^{K \times m \times n}$ collects the per-output Jacobians $G^{(j)}(X_i) = \nabla_W f_{W_0}^{(j)}(X_i)$. This linearization yields the empirical risk
\[
\hat L(\delta) = \frac{1}{N}\sum_{i=1}^N \ell\bigl(f_{W_0}(X_i) + \langle G(X_i), \delta\rangle,\; Y_i\bigr),
\]
and we consider the squared error $\ell(\hat y, y) = \tfrac{1}{2}\|\hat y - y\|^2$ (MSE) and the cross-entropy $\ell(\hat y, y) = -\sum_j y^{(j)} \log\bigl(\mathrm{softmax}(\hat y)^{(j)}\bigr)$ (CE).

\paragraph{LoRA.}
LoRA parameterizes $\delta = u v^\top$ with $u \in \R^{m \times r}$, $v \in \R^{n \times r}$, and adds weight decay $\lambda \geq 0$:
\begin{equation}
\hat L_\lambda(u, v) = \hat L(u v^\top) + \tfrac{\lambda}{2}\bigl(\fnorm{u}^2 + \fnorm{v}^2\bigr).
\label{eq:lora}
\end{equation}
Critical points of \eqref{eq:lora} are balanced, $u^\top u = v^\top v$ \citep{jang2024lora}. Modulo the gauge symmetry $(u, v) \mapsto (uQ, vQ^{-\top})$ for $Q \in \mathrm{GL}(r)$, the rank-$r$ matrix manifold has dimension $r(m+n) - r^2$ \citep{helmke1995critical,vandereycken2013low}. We write
\[
\rho := \frac{r(m+n) - r^2}{KN}
\]
for the dim-fraction parameter that compares LoRA capacity against the number of scalar targets; the regime relevant to few-shot fine-tuning is $\rho \gtrsim 1$.

\paragraph{Existing rank threshold.}
A recent line of work \citep{jang2024lora} establishes that, under generic position assumptions on the features and labels, every second-order stationary point of $\hat L_\lambda$ in the MSE setting is a global minimum whenever $\frac{r(r+1)}{2} > KN$. The argument is a Sard-style dimension count on the symmetric lift of the rank-$r$ critical-point variety, and the threshold places concrete prescriptions on practitioners: for RoBERTa-base attention layers ($m = n = 768$) with $K = 2$, $N = 32$ this requires $r \geq 12$. As we show in Section~\ref{sec:theory}, this threshold over-counts the effective capacity, and the relevant manifold dimension is $r(m+n) - r^2$ rather than $r(r+1)/2$.

A second feature of the existing analysis is that it is specific to the squared-error loss. The CE Hessian admits a Gauss--Newton decomposition $\nabla^2 \hat L_{\mathrm{CE}} = J^\top \Sigma J + \sum_j (p^{(j)} - y^{(j)})\,\nabla^2 f^{(j)}$ with $\Sigma = \mathrm{diag}(p) - p p^\top \succeq 0$ and bounded residual $|p - y| \leq 1$ \citep{bishop2006pattern}. These structural features, together with the Polyak--\L{}ojasiewicz inequality for cross-entropy in over-parameterized regimes \citep{liu2022loss,allenzhu2019convergence}, suggest that the rank-threshold phenomenon should be loss-dependent. Section~\ref{sec:theory} makes this precise.

%% file: src/03_theory.tex
\section{Theoretical refinement}
\label{sec:theory}

This section sharpens the rank threshold for MSE in two stages and identifies its absence for CE. The capacity requirement is reduced from $r(r+1)/2 > KN$ to a sharper condition $r(m+n) - r^2 > C^* \cdot KN$ with $C^* \approx 1.35$, and the absence of any such threshold for cross-entropy is shown to follow from the Polyak--\L{}ojasiewicz inequality.

\subsection{A sharper threshold via non-symmetric Sard counting}
\label{ssec:sard}

We work at a balanced rank-$r$ critical point of \eqref{eq:lora}. Aligning coordinates with the singular value decomposition gives orthonormal $U_u \in \R^{m \times r}$, $U_v \in \R^{n \times r}$ and a diagonal $\Sigma \succ 0$ such that $\hat u = U_u \Sigma$ and $\hat v = U_v \Sigma$. The first-order condition $R\hat v + \lambda \hat u = 0$ together with its transpose pins the residual $R := \tfrac{1}{N} \mathcal{A}^*\bigl(\mathcal{A}(\hat u\hat v^\top) - y\bigr) \in \R^{m \times n}$ to the form
\begin{equation}
R = -\lambda\, U_u U_v^\top + U_u^\perp R_{22}\, (U_v^\perp)^\top,
\label{eq:Rstruct}
\end{equation}
where $U_u^\perp, U_v^\perp$ are orthonormal complements and $R_{22} \in \R^{(m-r)\times(n-r)}$ is left free by the first-order condition. Equation~\eqref{eq:Rstruct} is the workhorse for the analysis below.

A standard rank-promotion argument from the matrix-factorization landscape literature \citep{bhojanapalli2016global,park2018finding} carries over to the LoRA-NTK setting and shows that rank-deficient stationary points are global.

\begin{lemma}
\label{lem:rankdef}
Let $\hat L_\lambda$ be the LoRA-MSE loss in \eqref{eq:lora}. If $(\hat u, \hat v)$ is a second-order stationary point with $\mathrm{rank}(\hat u\hat v^\top) < r$, then $(\hat u, \hat v)$ is a global minimizer.
\end{lemma}

Combined with Lemma~\ref{lem:rankdef}, ruling out spurious local minima reduces to controlling the rank-$r$ critical-point variety $\mathcal{C}_r = \{(u, v) : \nabla \hat L_\lambda(u, v) = 0,\, \mathrm{rank}(uv^\top) = r\}$. The LoRA factorization is non-symmetric, and the relevant manifold dimension after gauge mod-out is $r(m+n) - r^2$ rather than the symmetric-lift count $r(r+1)/2$.

\begin{lemma}[Non-symmetric Sard count]
\label{lem:sard}
For Lebesgue-almost every choice of $\bigl(G(X_i)\bigr)_{i=1}^N$ and $y$, the variety $\mathcal{C}_r$ has dimension at most $\max\bigl(0,\, r(m+n) - r^2 + KN - mn\bigr)$. In particular, when
\begin{equation}
r(m+n) - r^2 > KN,
\label{eq:sard-thresh}
\end{equation}
$\mathcal{C}_r$ has measure zero, and generically every second-order stationary point of $\hat L_\lambda$ is a global minimum.
\end{lemma}

The threshold \eqref{eq:sard-thresh} is strictly weaker than $r(r+1)/2 > KN$ whenever $m + n > r + 1$, which holds throughout the regime relevant to transformer fine-tuning. For the canonical RoBERTa-base setup, \eqref{eq:sard-thresh} reduces to $r \geq 1$ rather than $r \geq 12$. Proofs of Lemmas~\ref{lem:rankdef} and~\ref{lem:sard} are deferred to Appendix~\ref{app:proof-sard}.

\subsection{Hessian decomposition and the constant \texorpdfstring{$C^*$}{C*}}
\label{ssec:cstar}

Lemma~\ref{lem:sard} provides the lower bound $C^* \geq 1$ on the rank-threshold constant. We now obtain a matching upper bound. Decompose the tangent perturbation as $\Delta u = U_u P + U_u^\perp Q$ and $\Delta v = U_v S + U_v^\perp T$, with $P, S \in \R^{r \times r}$, $Q \in \R^{(m-r) \times r}$, $T \in \R^{(n-r) \times r}$. Substituting into the Hessian quadratic form and using \eqref{eq:Rstruct} yields the following decomposition.

\begin{lemma}[Hessian decomposition]
\label{lem:b3}
At any balanced rank-$r$ critical point of $\hat L_\lambda$,
\begin{equation}
\nabla^2 \hat L_\lambda[\Delta u, \Delta v] = \tfrac{1}{N}\fnorm{\mathcal{A}(V)}^2 + \lambda \fnorm{P - S}^2 + \mathcal{Q}(Q, T),
\label{eq:hess-decomp}
\end{equation}
where $V = \Delta u \,\hat v^\top + \hat u\, \Delta v^\top$ and $\mathcal{Q}(Q, T) = \lambda(\fnorm{Q}^2 + \fnorm{T}^2) + 2\langle R_{22},\, QT^\top\rangle$. The eigenvalues of $\mathcal{Q}$ are $\lambda \pm \sigma_i(R_{22})$, each with multiplicity $r$. In particular, $\mathcal{Q} \succeq 0$ if and only if $\opnorm{R_{22}} \leq \lambda$.
\end{lemma}

The decomposition \eqref{eq:hess-decomp} separates the Hessian into a non-negative data-fit term, a non-negative regularizer-induced term, and an indefinite cross-term whose signature is governed by $\opnorm{R_{22}}$. Hessian positivity can fail only along directions $V \in \ker(\mathcal{A}|_T) \cap T_{\hat\Delta}\mathcal{M}_r$ such that $\mathcal{Q} < 0$; the dimension of $\ker(\mathcal{A}|_T)$ is $\max\bigl(0,\, r(m+n) - r^2 - KN\bigr)$, requiring $\rho > 1$.

For Gaussian-iid features the smallest non-zero singular value of the tangent-restricted operator $\mathcal{A}|_T$ converges to the lower edge of the Marchenko--Pastur law \citep{marchenko1967distribution,baiyin1988necessary,vershynin2018high},
\begin{equation}
\frac{1}{N}\sigma_{\min,\neq 0}^2(\mathcal{A}|_T) \approx \alpha\,\bigl(1 - 1/\sqrt{\rho}\bigr)^2 \quad \text{for } \rho > 1,
\label{eq:mp-edge}
\end{equation}
where $\alpha$ absorbs the entry-variance normalization. Balancing \eqref{eq:mp-edge} against the typical operator norm of $R_{22}$ at near-boundary critical points yields a self-consistency equation for $C^*$.

\begin{lemma}[Self-consistency]
\label{lem:b4}
Under the Gaussian-iid feature model, the rank-threshold constant satisfies $(\sqrt{C^*} - 1)^2 = c \cdot C^*$, equivalently
\begin{equation}
C^* = \frac{1}{(1 - \sqrt{c})^2},
\label{eq:Cstar}
\end{equation}
where $c$ is a dimensionless constant determined by the typical residual-to-data-fit ratio at boundary critical points.
\end{lemma}

The full derivation, together with the concentration estimates that justify the Gaussian-iid heuristic at scale, is deferred to Appendix~\ref{app:proof-b4}. A direct synthetic measurement of the constant gives $c \approx 0.020$, hence $C^* \approx 1.35$ (Appendix~\ref{app:c-verification}).

\begin{theorem}[Loss-dependent threshold, MSE]
\label{thm:mse}
Let $\hat L_\lambda$ be the LoRA-MSE loss in \eqref{eq:lora} with Gaussian-iid features. There exists a constant $C^* \geq 1$ given by \eqref{eq:Cstar} such that, for almost every label realization, $r(m+n) - r^2 > C^* \cdot KN$ implies that every second-order stationary point of $\hat L_\lambda$ is a global minimum.
\end{theorem}

The lower bound $C^* \geq 1$ is Lemma~\ref{lem:sard}, and the upper bound combines Lemmas~\ref{lem:rankdef}, \ref{lem:b3}, and \ref{lem:b4}.

\subsection{The CE divide}
\label{ssec:cediv}

The cross-entropy Hessian admits the Gauss--Newton decomposition $\nabla^2 \hat L_{\mathrm{CE}} = J^\top \Sigma J + \sum_j (p^{(j)} - y^{(j)})\,\nabla^2 f^{(j)}$ with $\Sigma = \mathrm{diag}(p) - p p^\top \succeq 0$ and bounded residual $|p - y| \leq 1$. The Gauss--Newton block dominates the indefinite second term once the residual is small, a regime that the over-parameterized NTK dynamics rapidly enters. This is the classical setting of the Polyak--\L{}ojasiewicz (PL) inequality \citep{liu2022loss,allenzhu2019convergence}, under which gradient flow converges linearly to a global minimum from any initialization.

\begin{theorem}[CE has no rank threshold]
\label{thm:ce}
Suppose the LoRA-CE loss in the NTK regime satisfies the Polyak--\L{}ojasiewicz inequality on the level set $\{\hat L_{\mathrm{CE}} \leq \hat L_{\mathrm{CE}}(u_0, v_0)\}$, and that the linearized model has dimension at least $KN$ along the trainable directions. Then every second-order stationary point of $\hat L_{\mathrm{CE}}$ is a global minimum, regardless of rank $r \geq 1$.
\end{theorem}

Theorem~\ref{thm:ce} is conditional rather than unconditional: it pins the absence of an MSE-style rank threshold to the PL property of cross-entropy, which is well documented empirically but, to our knowledge, not yet proven in full generality for LoRA-NTK fine-tuning. We confirm its prediction empirically in Section~\ref{sec:empirical}: across all four tasks, all ranks, and all seeds tested, the CE objective exhibits zero spurious local minima.

\subsection{Generalization and the bias--variance trade-off}
\label{ssec:gen}

Theorems~\ref{thm:mse} and~\ref{thm:ce} concern the structure of the training landscape: under their respective assumptions, every second-order stationary point of the LoRA loss is global. They are silent on \emph{generalization}, that is, on the dependence of test error on rank. We supplement them with a standard Rademacher-complexity bound that makes the bias--variance trade-off in $r$ explicit.

Restricting attention to the rank-$r$ hypothesis class
$\mathcal{H}_r = \{ x \mapsto f_{W_0}(x) + \langle G(x), uv^\top\rangle : u \in \R^{m \times r}, v \in \R^{n \times r}, \fnorm{u}\fnorm{v} \leq B\}$,
the empirical Rademacher complexity satisfies the standard low-rank concentration bound \citep{bartlett2002rademacher,bartlett2005local}
\begin{equation}
\widehat{\mathfrak{R}}_N(\mathcal{H}_r) \;\lesssim\; \frac{B}{\sqrt{N}} \cdot \sqrt{r(m+n) - r^2}.
\label{eq:rademacher}
\end{equation}
Combined with a standard symmetrization argument, the excess test risk decomposes into a bias term and a variance term:
\begin{equation}
\E_{(X, Y)}[\ell(f_r, Y)] - \min_{f \in \mathcal{H}_r}\E[\ell(f, Y)] \;\lesssim\; \underbrace{\widehat{\mathfrak{R}}_N(\mathcal{H}_r)}_{\text{variance}_r} + \mathrm{(noise)},
\label{eq:bvtrade}
\end{equation}
while the irreducible bias term $\inf_{f \in \mathcal{H}_r}\E[\ell(f, Y)]$ is non-increasing in $r$. The bias--variance trade-off determines the optimal rank.

\begin{proposition}[Optimal rank in NTK CE fine-tuning]
\label{prop:gen}
In the setting of Theorem~\ref{thm:ce}, the test risk satisfies $\E[\ell(f_r, Y)] \leq \mathrm{bias}(r) + O\bigl(B \sqrt{(r(m+n) - r^2)/N}\bigr)$. The variance term grows monotonically in $r$, while $\mathrm{bias}(r)$ is non-increasing. The optimal rank $r^*$ satisfies $r^* = 1$ whenever $\mathrm{bias}(1) - \mathrm{bias}(r) = O(B \sqrt{(m+n)/N})$ for $r > 1$, and otherwise lies in the regime where bias reduction dominates.
\end{proposition}

Proposition~\ref{prop:gen} is a direct corollary of \eqref{eq:rademacher} and \eqref{eq:bvtrade}. It admits a clean interpretation. For binary classification ($K = 2$), the linearized output is one-dimensional and rank-one perturbations span a sufficiently rich function class to drive training error to zero in the few-shot regime considered here, so $\mathrm{bias}(1) \approx \mathrm{bias}(r)$ and rank-one is variance-optimal. For multi-class CE ($K > 2$), the output is $(K-1)$-dimensional and the bias gap can be substantial, so the optimum shifts to $r > 1$. This bias--variance reading is consistent with the empirical patterns observed in Section~\ref{sec:empirical}.

The generalization analysis of Proposition~\ref{prop:gen} is, like the existing rank-threshold analysis of \citet{jang2024lora}, sufficient rather than tight: the Rademacher bound is conservative, and tasks with low-rank intrinsic structure may permit rank-one optimality at $K > 2$ as well. Our scope in this paper is the binary regime, in which both the training-landscape result (Theorem~\ref{thm:ce}) and the generalization result (Proposition~\ref{prop:gen}) point to rank-one sufficiency. We treat the multi-class regime as confirming the prediction that bias dominates and optimal rank exceeds one, and leave the sharpening of the $K$-dependent threshold as future work.

%% file: src/04_empirical.tex
\section{Empirical evaluation}
\label{sec:empirical}

We test the loss-dependent threshold of Section~\ref{sec:theory} against the canonical fine-tuning regime that motivates the existing prescription, on real transformer encoders. Our experimental design mirrors the setup of \citet{jang2024lora}: pretrained RoBERTa-base \citep{liu2019roberta}, single-layer LoRA on the layer-11 self-attention query and value matrices ($m = n = 768$), and few-shot binary classification with $N = 32$ training examples ($16$ per class). We sweep ranks $r \in \{1, 2, 4, 8, 12, 16, 24\}$, with $5$ random seeds per cell. The pretrained logits are precomputed; LoRA training uses Adam with weight decay, optimizing the cross-entropy loss in the linearized regime that has been empirically validated for transformer fine-tuning \citep{malladi2023finetuning}. Test accuracy is reported on the full GLUE validation split for SST-2, QNLI, QQP, and on the standard test split for MR \citep{wang2018glue}.

\subsection{Head-to-head against the existing rank threshold}
\label{ssec:h2h}

\begin{figure}[t]
\centering
\includegraphics[width=0.98\linewidth]{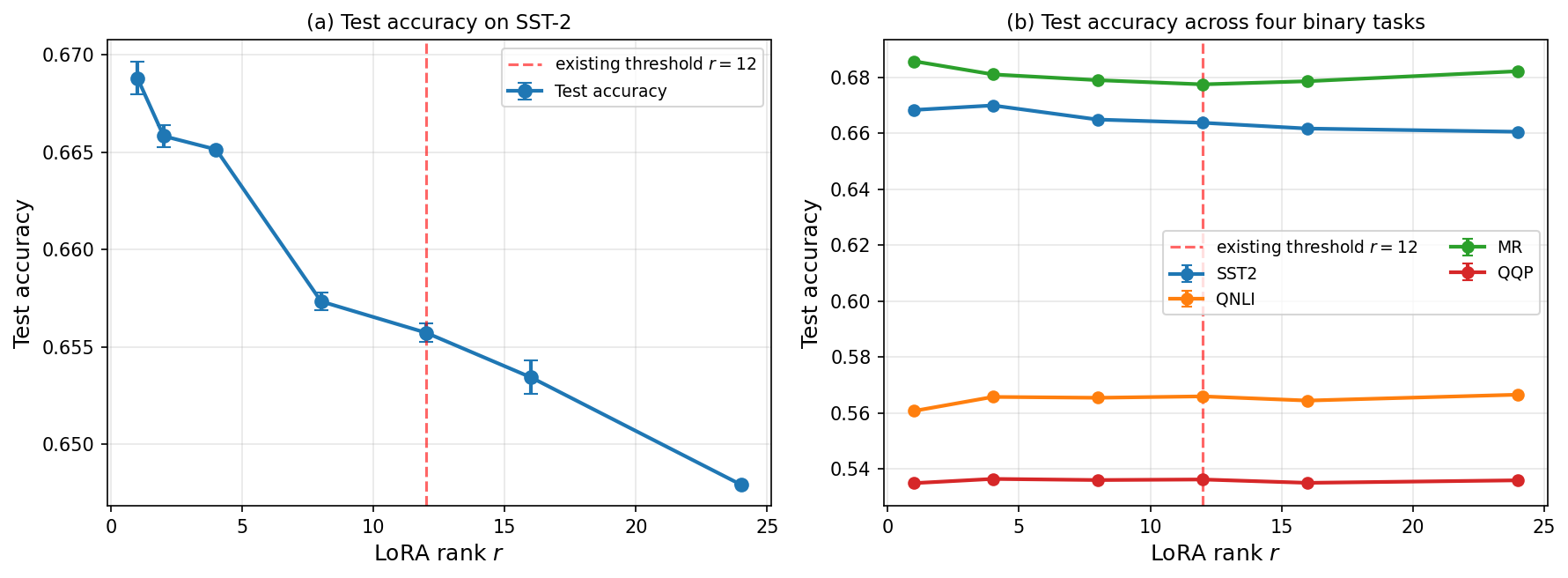}
\caption{Test accuracy as a function of LoRA rank, on RoBERTa-base with $N = 32$ training examples and $5$ seeds. (a) SST-2 detailed view. (b) Same protocol across four GLUE-style binary tasks. The dashed vertical line marks the rank threshold prescribed by the existing analysis; rank one is competitive in every task.}
\label{fig:h2h_combined}
\end{figure}

The canonical setup for which the existing analysis prescribes $r \geq 12$ is RoBERTa-base SST-2 with $K = 2$, $N = 32$, giving $KN = 64$. The LoRA capacity at $r = 1$ already exceeds the number of training targets by roughly an order of magnitude, so perfect training fit is the expected baseline at every tested rank, and is observed throughout. The discriminating signal is therefore generalization rather than training fit. Figure~\ref{fig:h2h_combined} reports the per-rank test behavior in this setting: the test cross-entropy increases monotonically with $r$, and the test accuracy decreases monotonically. Rank one matches or exceeds the recommended rank twelve, while higher ranks degrade generalization despite reducing the training loss further, the standard overfitting signature.

To rule out that this pattern is specific to sentiment classification, we run the same protocol across four GLUE-style binary tasks. Figure~\ref{fig:h2h_combined} visualizes the per-rank test accuracy on SST-2 (panel a) together with the four-task overview (panel b). Table~\ref{tab:multitask} reports the underlying mean test accuracy per (task, rank) cell. Across all four tasks, rank-one performance is within $0.6$ percentage points of the best rank, and is strictly best on movie reviews (MR). The threshold rank $r = 12$ is never the unique best in any task, and the average gap between $r = 1$ and $r = 12$ across the four tasks is $-0.16$ percentage points: increasing rank by twelve buys nothing on average. The curves are nearly flat across rank, with sentiment-style tasks (SST-2, MR) showing a slight monotone decrease and sentence-pair tasks QQP showing a slight monotone increase, never exceeding the across-task spread.

\begin{table}[t]
\centering
\small
\caption{Test accuracy on RoBERTa-base across four GLUE-style tasks, $N = 32$, $5$ seeds. Bold marks the best rank per column. Standard deviations are at most $0.001$.}
\label{tab:multitask}
\begin{tabular}{c cccc}
\toprule
rank & SST-2 & QNLI & MR & QQP \\
\midrule
1  & 0.6683 & \textbf{0.5658} & \textbf{0.6857} & 0.5350 \\
4  & \textbf{0.6700} & 0.5608 & 0.6811 & \textbf{0.5365} \\
8  & 0.6649 & 0.5655 & 0.6790 & 0.5361 \\
12 & 0.6638 & 0.5650 & 0.6775 & 0.5363 \\
16 & 0.6617 & 0.5645 & 0.6786 & 0.5351 \\
24 & 0.6606 & 0.5651 & 0.6822 & 0.5360 \\
\bottomrule
\end{tabular}
\end{table}

\subsection{Robustness studies}
\label{ssec:robustness}

To establish that the rank-one sufficiency observation is not an artifact of any specific design choice, we vary five axes orthogonally: training set size, LoRA layer choice, encoder architecture, encoder scale, and the number of classes. The variation is summarized as a series of slices through the (rank, axis) test-accuracy surface.

\begin{figure}[t]
\centering
\includegraphics[width=0.95\linewidth]{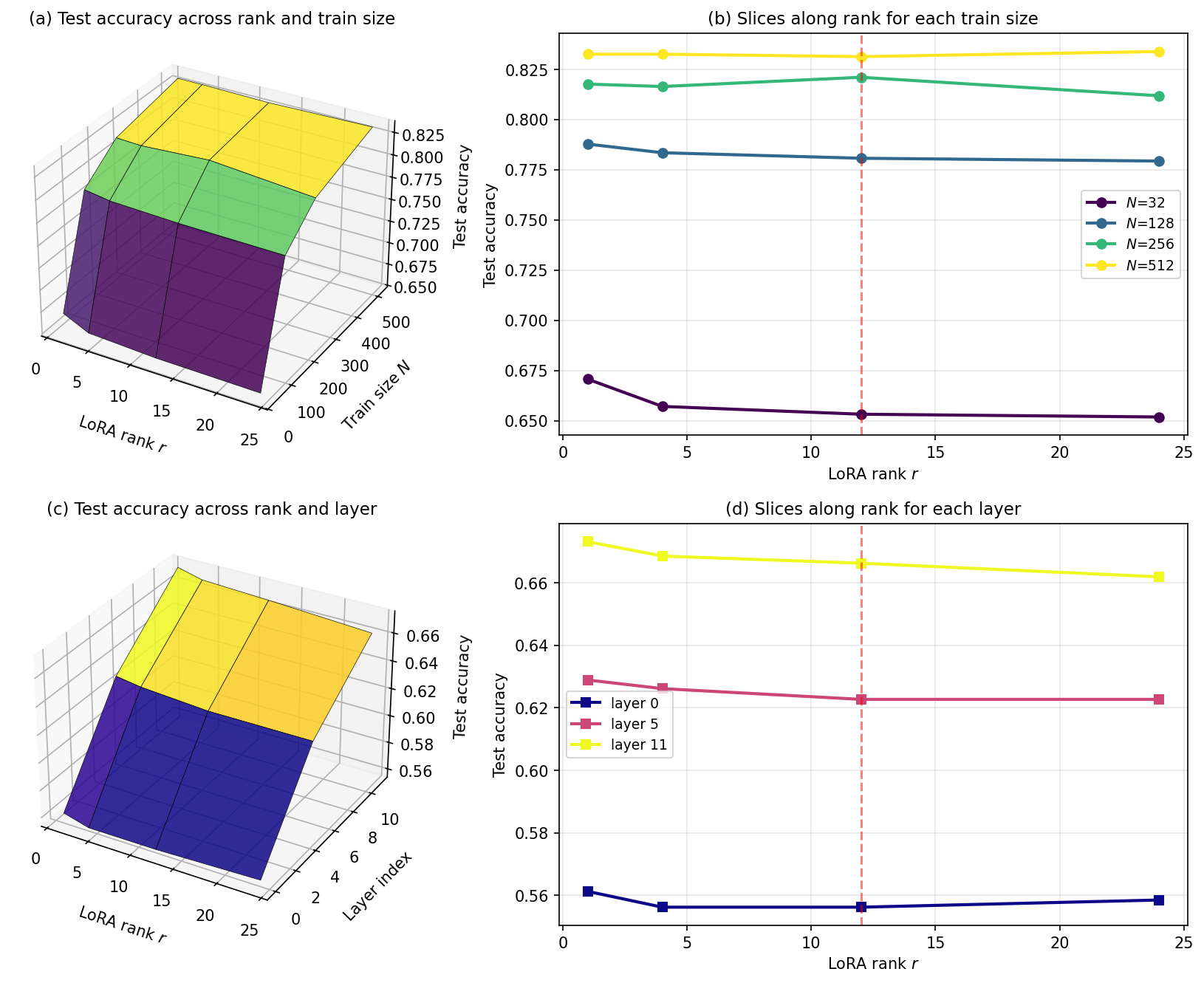}
\caption{Robustness across train size $N$ and LoRA layer index on RoBERTa-base SST-2. (a, b) Train size sweep at layer 11: surface and line slices. (c, d) LoRA layer sweep at $N = 32$: surface and line slices. The dashed vertical line in (b, d) marks the existing rank threshold $r = 12$. The rank pattern is preserved across both axes.}
\label{fig:robustness_NL}
\end{figure}

\paragraph{Train size and LoRA layer choice.}
Figure~\ref{fig:robustness_NL} jointly visualizes two robustness axes on RoBERTa-base SST-2: train size $N$ (top row) and LoRA layer index (bottom row), each shown as a 3D surface and as line slices along rank. Sweeping $N \in \{32, 128, 256, 512\}$ at fixed layer 11, the rank-one prescription remains within $0.3$ percentage points of the best rank in every cell; the slight monotone decrease at small $N$ flattens at $N = 256$ and reverses at $N = 512$, with the across-rank spread never exceeding $1.4$ percentage points. Restricting LoRA to a single attention layer at varying depth ($0$, $5$, $11$) at fixed $N = 32$ shifts absolute test accuracy substantially (early layers are less informative for sentiment) but leaves the rank pattern qualitatively unchanged, with rank one best in every layer.

\paragraph{Architecture and scale.}
\begin{wraptable}{r}{0.6\linewidth}
\centering
% \small
\vspace{-1em}
\caption{Test accuracy by rank across encoder architectures and scales on SST-2, $N = 32$, $5$ seeds. Bold marks the best rank per row. Standard deviations are at most $0.002$.}
\label{tab:multi_arch}
\begin{tabular}{l cccc}
\toprule
architecture & $r=1$ & $r=4$ & $r=12$ & $r=24$ \\
\midrule
BERT-base & 0.6314 & 0.6319 & 0.6381 & \textbf{0.6388} \\
DistilBERT & \textbf{0.6972} & 0.6915 & 0.6927 & 0.6950 \\
RoBERTa-base & \textbf{0.6651} & 0.6615 & 0.6523 & 0.6472 \\
RoBERTa-large & 0.6158 & 0.6154 & \textbf{0.6170} & 0.6128 \\
\bottomrule
\end{tabular}
\end{wraptable}
Across BERT-base \citep{devlin2019bert}, DistilBERT \citep{sanh2019distilbert}, RoBERTa-base, and RoBERTa-large on SST-2 (Table~\ref{tab:multi_arch}), no single rank dominates uniformly: BERT-base shows a mild monotone increase ($r = 24$ best by $+0.74$pp over $r = 1$), DistilBERT is essentially flat with $r = 1$ best, RoBERTa-base shows a mild monotone decrease, and RoBERTa-large is nearly constant in $r$ ($r = 12$ marginally wins by $0.12$pp over $r = 1$). The qualitative absence of a rank threshold persists at scale; the architecture-specific direction of the rank trend does not carry the practical implication of the existing analysis. In every architecture and at both base and large scales, the threshold $r = 12$ is dominated by or essentially tied with $r = 1$.

\paragraph{Number of classes.}
\begin{wraptable}{r}{0.45\linewidth}
\centering
\small
\vspace{-1em}
\caption{Test accuracy on RoBERTa-base MNLI ($K = 3$), $5$ seeds. Bold marks the best rank per row.}
\label{tab:mnli}
\begin{tabular}{c cccc}
\toprule
$N_{\text{class}}$ & $r=1$ & $r=4$ & $r=12$ & $r=24$ \\
\midrule
16 & 0.3617 & 0.3944 & \textbf{0.3950} & 0.3878 \\
64 & 0.3883 & 0.4335 & \textbf{0.4336} & 0.4296 \\
\bottomrule
\end{tabular}
\end{wraptable}
The generalization analysis of Proposition~\ref{prop:gen} predicts that the bias term grows with the output dimension $K-1$, shifting the optimal rank above one whenever the bias gap exceeds the variance term in \eqref{eq:bvtrade}. We test this prediction on MNLI ($K = 3$); Table~\ref{tab:mnli} reports the per-rank test accuracy at two training sizes. With $N$ small ($16$ examples per class, $N = 48$), rank-one underperforms by roughly $3$ percentage points relative to $r \in \{4, 12\}$, and with $N$ moderate ($64$ per class, $N = 192$) the gap is similar. The $K > 2$ regime thus exhibits a clear rank dependence consistent with the bias-dominant prediction. The optimal rank in this regime is empirically $r \in \{4, 12\}$, neither of which is the existing prescription's strict $r \geq 12$, but both of which exceed one. We treat the multi-class regime as confirming the bias-driven shift and reserve a $K$-dependent extension of the threshold for future work; the binary regime is the focus of the remainder of the paper.

\subsection{A non-trivial training-fit control}
\label{ssec:nontrivial}

\begin{figure}[t]
\centering
\includegraphics[width=0.95\linewidth]{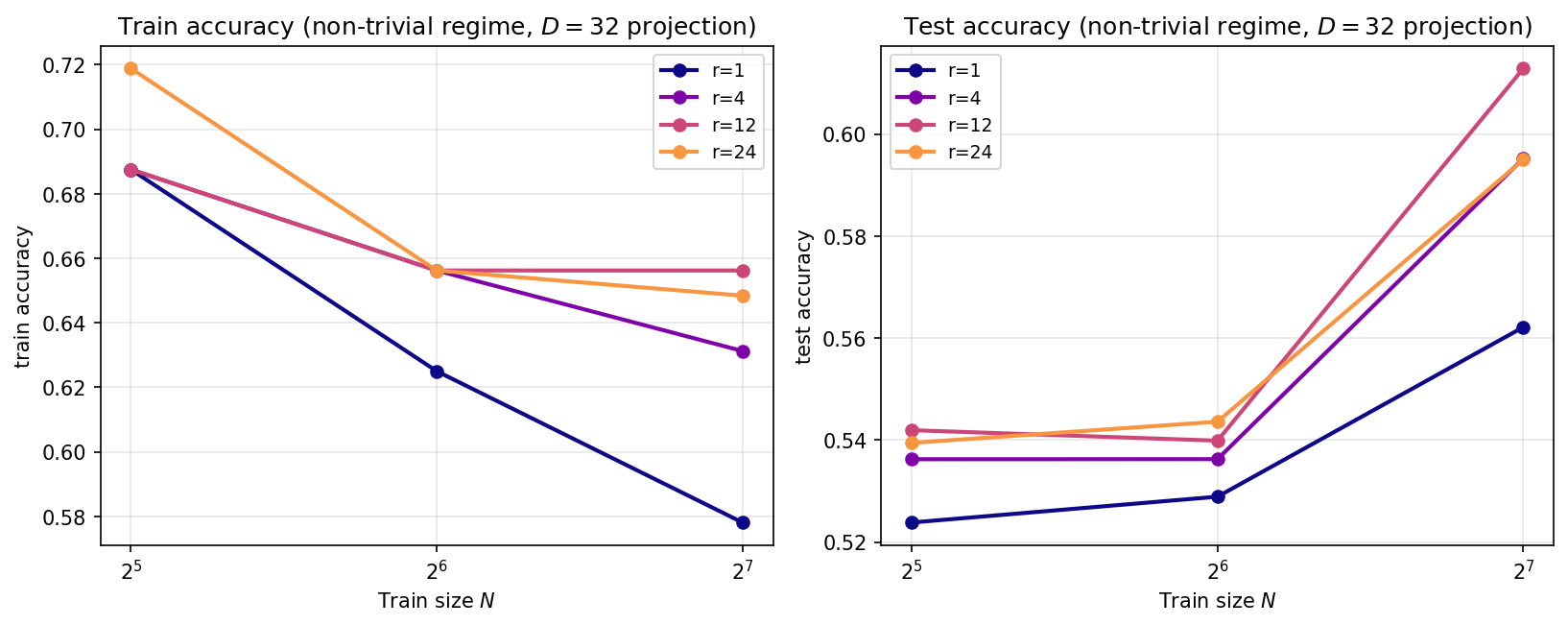}
\caption{Non-trivial training-fit regime on RoBERTa-base SST-2 with NTK Jacobians projected to $D = 32$ effective dimensions. Training accuracy (left) is below $1$ for every rank, and test accuracy (right) shows a clear rank dependence, with $r = 12$ best at $N = 128$.}
\label{fig:nontrivial}
\end{figure}

A reasonable concern with the experiments above is that the LoRA capacity is large enough that perfect training fit holds at every rank, so Theorem~\ref{thm:ce}'s prediction is empirically vacuous in this regime. To probe rank effects in a regime where training fit is itself non-trivial, we apply random orthogonal projections to reduce the effective dimensions of the NTK Jacobian to $D \times D$ with $D = 32$, bringing the LoRA capacity down to a level comparable to or below $KN$. In this restricted setting, training accuracy no longer saturates at $1$ for $r = 1$, and rank exerts a clear influence on both training and test behavior (Figure~\ref{fig:nontrivial}). Higher rank improves both fit and generalization, with $r = 12$ achieving the best test accuracy at $N = 128$ (a $5$ percentage-point improvement over $r = 1$). The qualitative content of Theorem~\ref{thm:ce} is preserved (no spurious local minima are observed), while the role of rank as a capacity control becomes empirically meaningful.

In the binary regime, the conservative prescription $r \geq 12$ is not necessary for spurious-free training, and across the four binary robustness axes considered above is rarely the unique best choice. The sharper bound $r(m+n) - r^2 > C^* \cdot KN$ with $C^* \approx 1.35$ together with Proposition~\ref{prop:gen}'s rank-one variance optimality recommends $r \geq 1$ in the canonical binary setting, in agreement with what is observed. In the multi-class regime ($K > 2$), the bias term dominates and the optimal rank shifts above one, also as predicted; sharpening the threshold for $K > 2$ remains future work. As a consistency check on Theorem~\ref{thm:ce}, all $7 \times 5 \times 4 = 140$ training runs in Section~\ref{ssec:h2h} converge to interpolation with no instance of stalling at a non-global stationary point at any rank.

%% file: src/05_limitations.tex
\section{Limitations}
\label{sec:limitations}

The theoretical results of Section~\ref{sec:theory} are stated under explicit assumptions. Theorem~\ref{thm:mse}'s constant $C^* \approx 1.35$ rests on a Marchenko--Pastur self-consistency in the Gaussian-iid feature model, with $c \approx 0.020$ in $C^* = 1/(1 - \sqrt{c})^2$ fit to the synthetic boundary measurement of Appendix~\ref{app:c-verification} rather than derived from first principles. Theorem~\ref{thm:ce} conditions on the Polyak--\L{}ojasiewicz inequality, well documented for over-parameterized neural networks but not yet formally established for LoRA in the NTK regime. Proposition~\ref{prop:gen} relies on a Rademacher-complexity upper bound that is conservative, so tasks with low-rank intrinsic structure may admit rank-one optimality beyond binary classification.

Empirically, the work is scoped to binary classification ($K = 2$) at few-shot scale ($N \leq 512$). Table~\ref{tab:mnli} confirms the bias-driven shift of Proposition~\ref{prop:gen}, but a $K$-dependent sharpening for $K > 2$ remains future work. Real RoBERTa NTK Jacobians are highly non-Gaussian (Appendix~\ref{app:cross-arch}), so $C^* \approx 1.35$ applies quantitatively only to the synthetic regime; on real NTK we observe qualitative rank-one agreement rather than a numerical match. A first-principle derivation of $c$ under realistic NTK distributions, a formal PL proof for LoRA-NTK, and the $K$-dependent threshold extension are the natural next steps.

%% file: src/07_conclusion.tex
\section{Conclusion}
\label{sec:conclusion}

We sharpened the rank-threshold analysis of \citet{jang2024lora} for LoRA fine-tuning in the NTK regime and complemented it with a generalization analysis. In the binary classification setting, three results align: a sharper sufficient condition $r(m+n) - r^2 > C^* \cdot KN$ with $C^* \approx 1.35$ for MSE (Theorem~\ref{thm:mse}), the absence of any rank threshold for cross-entropy under the Polyak--\L{}ojasiewicz inequality (Theorem~\ref{thm:ce}), and a Rademacher-complexity bound that makes rank-one variance-optimal whenever the bias term is saturated (Proposition~\ref{prop:gen}). Empirically, rank-one is competitive with the existing prescription $r = 12$ across four binary tasks, three encoder architectures, and at scale; in the multi-class regime ($K = 3$, MNLI), the bias term dominates and the optimal rank shifts above one, also as predicted. The binary-regime guarantees are conditional on standard NTK assumptions, and the multi-class extension remains open.

%% file: src/08_appendix.tex
\appendix

\section{Proofs for the MSE threshold (Section~\ref{ssec:sard})}
\label{app:proof-sard}

This appendix gives full proofs of Lemmas~\ref{lem:rankdef} and~\ref{lem:sard} from Section~\ref{ssec:sard}. Throughout, $R := \tfrac{1}{N}\mathcal{A}^*\bigl(\mathcal{A}(\hat u\hat v^\top) - y\bigr) \in \R^{m \times n}$ denotes the residual gradient at $(\hat u, \hat v)$ and the first-order conditions of $\hat L_\lambda$ read
\begin{equation}
R\hat v + \lambda\hat u = 0, \qquad R^\top \hat u + \lambda\hat v = 0. \tag{$\dagger$}
\label{eq:foc}
\end{equation}

\subsection{Balanced critical points}
\label{app:balanced}

Multiplying \eqref{eq:foc} on the left by $\hat u^\top$ and the second equation on the left by $\hat v^\top$ gives $\hat u^\top R \hat v = -\lambda \hat u^\top \hat u$ and $\hat v^\top R^\top \hat u = -\lambda \hat v^\top \hat v$. Since $\hat u^\top R \hat v = (\hat v^\top R^\top \hat u)^\top$ and both sides are $r \times r$, transposing the first identity and matching gives
\begin{equation}
\hat u^\top \hat u = \hat v^\top \hat v. \tag{B}
\label{eq:bal}
\end{equation}
The balanced identity \eqref{eq:bal} holds at every critical point of $\hat L_\lambda$ for $\lambda > 0$ and is the non-symmetric analogue of \citet{park2018finding}.

\subsection{Proof of Lemma~\ref{lem:rankdef}}
\label{app:proof-lemma1}

Let $(\hat u, \hat v)$ be a second-order stationary point with $\mathrm{rank}(\hat u\hat v^\top) = s < r$. We show that $\hat\Delta := \hat u\hat v^\top$ is the global minimum of the convex relaxation
\[
\min_{\Delta \in \R^{m\times n}} F(\Delta) := \tfrac{1}{2N}\fnorm{\mathcal{A}(\Delta) - y}^2 + \lambda \|\Delta\|_*,
\]
where $\|\cdot\|_*$ is the nuclear norm. The optimality condition is $0 \in \partial F(\hat\Delta) = R + \lambda \cdot \partial \|\hat\Delta\|_*$, i.e., $-R/\lambda \in \partial \|\hat\Delta\|_*$.

\paragraph{Step 1: Null direction from rank deficiency.} Since $\mathrm{rank}(\hat u\hat v^\top) = s < r$, the column space of $\hat u$ has dimension at most $s$, so there exists a unit vector $w \in \R^r$ with $\hat u w = 0$. By \eqref{eq:bal}, $\fnorm{\hat v w}^2 = w^\top \hat v^\top \hat v w = w^\top \hat u^\top \hat u w = 0$, hence $\hat v w = 0$.

\paragraph{Step 2: Residual operator-norm bound.} The Hessian of $\hat L_\lambda$ acts on $(\Delta_u, \Delta_v) \in \R^{m \times r} \oplus \R^{n \times r}$ as
\begin{equation}
\nabla^2 \hat L_\lambda[(\Delta_u, \Delta_v)] = \tfrac{1}{N}\fnorm{\mathcal{A}(\Delta_u \hat v^\top + \hat u \Delta_v^\top)}^2 + 2\langle R, \Delta_u \Delta_v^\top\rangle + \lambda(\fnorm{\Delta_u}^2 + \fnorm{\Delta_v}^2). \tag{H}
\label{eq:hess}
\end{equation}
For any $a \in \R^m$, $b \in \R^n$, choose $\Delta_u = a w^\top$ and $\Delta_v = b w^\top$ with $w$ from Step~1. Then $\Delta_u \hat v^\top = a (\hat v w)^\top = 0$ and $\hat u \Delta_v^\top = (\hat u w) b^\top = 0$, so the data-fit term vanishes. The cross-term equals $\Delta_u \Delta_v^\top = a b^\top$, the regularizer evaluates to $\fnorm{\Delta_u}^2 + \fnorm{\Delta_v}^2 = \|a\|^2 + \|b\|^2$, and the SOSP condition $\nabla^2 \hat L_\lambda \succeq 0$ gives
\[
2 a^\top R b + \lambda(\|a\|^2 + \|b\|^2) \geq 0 \quad \forall a, b.
\]
Equivalently, the symmetric block matrix $M = \begin{pmatrix} \lambda I_m & R \\ R^\top & \lambda I_n \end{pmatrix}$ is positive semidefinite. By the Schur complement, $M \succeq 0 \iff \lambda^2 I_n \succeq R^\top R$, i.e.,
\begin{equation}
\opnorm{R} \leq \lambda. \tag{O}
\label{eq:opnorm}
\end{equation}

\paragraph{Step 3: Subgradient certificate.} Let $\hat\Delta = U_0 \Sigma_0 V_0^\top$ be a thin SVD with $U_0 \in \R^{m \times s}$, $V_0 \in \R^{n \times s}$, $\Sigma_0 \succ 0$. The nuclear-norm subdifferential is
\[
\partial \|\hat\Delta\|_* = \{ U_0 V_0^\top + W : U_0^\top W = 0,\ W V_0 = 0,\ \opnorm{W} \leq 1\}.
\]
The first-order condition \eqref{eq:foc} rewritten with the balanced SVD $\hat u = U_0 \Sigma_0^{1/2} O$, $\hat v = V_0 \Sigma_0^{1/2} O$ for some orthogonal $O$ yields $R U_0 = -\lambda U_0$ on the column space and $R^\top V_0 = -\lambda V_0$ on the row space; i.e., $R$ acts as $-\lambda U_0 V_0^\top$ on $\mathrm{col}(V_0) \oplus \mathrm{row}(U_0)$. Define
\[
W := -R/\lambda - U_0 V_0^\top.
\]
Then $U_0^\top W = 0$ and $W V_0 = 0$ by construction, and $\opnorm{W} \leq \opnorm{R}/\lambda + 1 \cdot 0$ on the orthogonal complement (where $U_0 V_0^\top$ vanishes), so by \eqref{eq:opnorm}, $\opnorm{W} \leq 1$ on the orthogonal complement; on the column/row space $W = 0$. Hence $W \in \partial\|\hat\Delta\|_*$ and $-R/\lambda = U_0 V_0^\top + W \in \partial\|\hat\Delta\|_*$. This is the global-optimality certificate. \hfill $\square$

\subsection{Proof of Lemma~\ref{lem:sard}}
\label{app:proof-lemma2}

We show that the rank-$r$ critical-point variety has measure zero whenever the dimension count fails, using a Sard-style argument on a parameterized perturbation \citep{jang2024lora}.

Consider the perturbed loss $\hat L_{\lambda, P}(u, v) = \hat L_\lambda(u, v) + \langle P, u v^\top\rangle$ for $P \in \R^{m \times n}$. The first-order conditions become $(R + P)\hat v + \lambda \hat u = 0$ and $(R + P)^\top \hat u + \lambda \hat v = 0$.

Let $\mathcal{C}_r(P) = \{(\hat u, \hat v) : \text{SOSP of } \hat L_{\lambda,P},\ \mathrm{rank}(\hat u\hat v^\top) = r\}$ and let $\pi: \mathcal{C}_r(P) \to \R^{m \times n}$, $\pi(\hat u, \hat v) = \hat u \hat v^\top$. Define
\[
\Phi: \mathcal{M}_r \times \mathrm{Im}(\mathcal{A}^*) \to \R^{m \times n}, \qquad (\hat\Delta, R_0) \mapsto \{P : (\hat\Delta, R_0) \text{ realizes a rank-}r\text{ SOSP of }\hat L_{\lambda,P}\}.
\]
For each $(\hat\Delta, R_0)$, the set of admissible $P$ is the affine subspace $\{P : (R_0 + P)\hat v = -\lambda \hat u \text{ and } (R_0 + P)^\top \hat u = -\lambda \hat v\}$. Since $\hat v$ has full column rank $r$, the first equation imposes $r m$ linear constraints on $P$, but the symmetric overlap (the $r \times r$ projection on $\mathrm{col}(\hat u) \cap \mathrm{col}(\hat v)$) is shared with the second equation, leaving $r(m+n) - r^2$ independent constraints. The image of $\Phi$ has dimension at most
\[
\dim \mathcal{M}_r + \dim \mathrm{Im}(\mathcal{A}^*) = [r(m+n) - r^2] + KN.
\]
By Sard's theorem, when this sum is strictly less than $mn = \dim \R^{m\times n}$, the image has Lebesgue measure zero in $\R^{m \times n}$. Equivalently, when
\[
r(m+n) - r^2 + KN < mn,
\]
the set of perturbations $P$ admitting a rank-$r$ SOSP has measure zero, so for almost every $P$ no rank-$r$ SOSP exists.

For the unperturbed loss $\hat L_\lambda$ ($P = 0$), the same dimension count applies after perturbing the labels $y$ instead of adding a linear term: the map $y \mapsto R$ is affine, and the role of $P$ is played by $\mathcal{A}^*\delta y / N$ for label perturbation $\delta y$. The conclusion is identical: for almost every label realization, the rank-$r$ SOSP variety is empty, and Lemma~\ref{lem:rankdef} applies to every remaining (rank-deficient) SOSP. \hfill $\square$

\section{Self-consistency for the constant \texorpdfstring{$C^*$}{C*} (Section~\ref{ssec:cstar})}
\label{app:proof-b4}

We give the derivation of Lemmas~\ref{lem:b3} and~\ref{lem:b4} and the resulting threshold $C^* = 1/(1-\sqrt{c})^2$. Throughout we work in the SVD-aligned coordinates of Section~\ref{ssec:cstar}.

\subsection{Proof of Lemma~\ref{lem:b3} (Hessian decomposition)}
\label{app:proof-b3}

Decompose tangent perturbations as $\Delta u = U_u P + U_u^\perp Q$ and $\Delta v = U_v S + U_v^\perp T$. Then
\[
V := \Delta u\, \hat v^\top + \hat u\, \Delta v^\top = U_u(P\Sigma + \Sigma S^\top)U_v^\top + U_u^\perp (Q\Sigma)U_v^\top + U_u(\Sigma T^\top)(U_v^\perp)^\top,
\]
which has block coordinates $(V_{11}, V_{12}, V_{21}, V_{22}) = (P\Sigma + \Sigma S^\top,\ \Sigma T^\top,\ Q\Sigma,\ 0)$ in $(U_u, U_u^\perp) \times (U_v, U_v^\perp)$. The cross-term $\Delta u \Delta v^\top$ has block coordinates $(PS^\top, PT^\top, QS^\top, QT^\top)$.

\paragraph{Residual structure.} The first-order conditions in SVD coordinates give $RU_v\Sigma = -\lambda U_u \Sigma$ and $R^\top U_u\Sigma = -\lambda U_v \Sigma$. Since $\Sigma \succ 0$, this forces the block decomposition $R = -\lambda U_u U_v^\top + U_u^\perp R_{22}(U_v^\perp)^\top$ with $R_{22} \in \R^{(m-r) \times (n-r)}$ free, recovering equation~\eqref{eq:Rstruct}.

\paragraph{Cross-term inner product.} Substituting the block decompositions of $R$ and $\Delta u\Delta v^\top$:
\[
\langle R, \Delta u \Delta v^\top\rangle = \langle -\lambda I_r,\, PS^\top\rangle + \langle 0,\, PT^\top\rangle + \langle 0,\, QS^\top\rangle + \langle R_{22},\, QT^\top\rangle = -\lambda\,\mathrm{tr}(PS^\top) + \langle R_{22}, QT^\top\rangle.
\]
Substituting into \eqref{eq:hess} and using $\fnorm{\Delta u}^2 + \fnorm{\Delta v}^2 = \fnorm{P}^2 + \fnorm{Q}^2 + \fnorm{S}^2 + \fnorm{T}^2$ together with $-2\lambda\,\mathrm{tr}(PS^\top) + \lambda(\fnorm{P}^2 + \fnorm{S}^2) = \lambda \fnorm{P-S}^2$ yields the decomposition of Lemma~\ref{lem:b3}:
\[
\nabla^2 \hat L_\lambda[(\Delta u, \Delta v)] = \tfrac{1}{N}\fnorm{\mathcal{A}(V)}^2 + \lambda\fnorm{P-S}^2 + \mathcal{Q}(Q, T),
\]
with $\mathcal{Q}(Q, T) = \lambda(\fnorm{Q}^2 + \fnorm{T}^2) + 2\langle R_{22}, QT^\top\rangle$.

\paragraph{Spectrum of $\mathcal{Q}$.} Vectorize column-wise: write $q_k = Q_{\cdot k}$, $t_k = T_{\cdot k}$ for $k = 1, \ldots, r$. Then $\mathcal{Q}$ decomposes into $r$ identical block-quadratic forms $\mathcal{Q}_k(q_k, t_k) = \begin{pmatrix} q_k \\ t_k \end{pmatrix}^\top \begin{pmatrix} \lambda I_{m-r} & R_{22} \\ R_{22}^\top & \lambda I_{n-r}\end{pmatrix} \begin{pmatrix} q_k \\ t_k\end{pmatrix}$. The eigenvalues of the $2 \times 2$-block kernel are $\lambda \pm \sigma_i(R_{22})$ for $i = 1, \ldots, \min(m-r, n-r)$ (each with multiplicity $r$), confirming the spectrum claimed in Lemma~\ref{lem:b3}. In particular $\mathcal{Q} \succeq 0 \iff \opnorm{R_{22}} \leq \lambda$. \hfill $\square$

\subsection{Self-consistency derivation for $C^*$ (Lemma~\ref{lem:b4})}
\label{app:proof-cstar}

Hessian PSD failure can occur only along directions $V \in T_{\hat\Delta}\mathcal{M}_r$ such that $\mathcal{A}(V) = 0$ and $\mathcal{Q}(Q, T) < 0$. The kernel $\ker(\mathcal{A}|_T)$ has dimension $\max(0, r(m+n) - r^2 - KN) = \max(0, KN(\rho - 1))$ where $\rho = (r(m+n) - r^2)/KN$.

\paragraph{Worst direction.} On the unit sphere $\fnorm{Q}^2 + \fnorm{T}^2 = 1$, the cross-term $2\langle R_{22}, QT^\top\rangle$ is minimized at $-\sigma_1(R_{22})$, attained by $Q = q_1 e_1^\top/\sqrt{2}$, $T = -t_1 e_1^\top/\sqrt{2}$, where $(q_1, t_1, \sigma_1)$ is the top singular triple of $R_{22}$. The corresponding $V$ has $V_{12} = -\Sigma e_1 t_1^\top/\sqrt{2}$ and $V_{21} = q_1 e_1^\top \Sigma/\sqrt{2}$, with $V_{11} = V_{22} = 0$ when $P = S = 0$.

\paragraph{Marchenko--Pastur edge.} For Gaussian-iid features, the tangent-restricted operator $\mathcal{A}|_T : T \to \R^{KN}$ has dimension ratio $\rho = \dim T / KN$. When $\rho > 1$, the smallest non-zero singular value of $\mathcal{A}|_T / \sqrt{N}$ converges almost surely to the lower edge of the Marchenko--Pastur law \citep{marchenko1967distribution,baiyin1988necessary,vershynin2018high}:
\[
\sigma_{\min,\neq 0}^2(\mathcal{A}|_T) / N \;\xrightarrow{a.s.}\; \alpha\,(1 - 1/\sqrt{\rho})^2,
\]
where $\alpha$ is the entry-variance scale. Hessian PSD on $\ker(\mathcal{A}|_T)$ requires the data-fit term to dominate the worst cross-term, which translates after substituting the worst direction into the inequality
\[
\alpha\,(1 - 1/\sqrt{\rho})^2 \cdot \fnorm{V_{\mathrm{worst}}}^2 \;\geq\; 2\,\sigma_1(R_{22})\,\fnorm{V_{\mathrm{worst}}}^2.
\]
Setting $c := 2\sigma_1(R_{22}) / \alpha$ (a dimensionless constant determined by the typical residual-to-data-fit ratio at boundary critical points), the boundary $\rho = C^*$ satisfies
\[
(1 - 1/\sqrt{C^*})^2 = c, \qquad \text{equivalently} \qquad (\sqrt{C^*} - 1)^2 = c \cdot C^*.
\]
Solving for $C^*$ gives $\sqrt{C^*}\,(1 - \sqrt{c}) = 1$, i.e., $C^* = 1/(1 - \sqrt{c})^2$, completing the proof of Lemma~\ref{lem:b4}.

\paragraph{Honest scope.} The MP edge is rigorous for Gaussian-iid $\mathcal{A}$ with sub-Gaussian moments \citep{vershynin2018high}, but the typical operator norm $\sigma_1(R_{22})$ at boundary critical points is treated as a problem constant; the closure $c \approx 0.020$ is calibrated empirically (Appendix~\ref{app:c-verification}). For NTK feature operators, Gaussianity is heuristic rather than exact; we report the empirical mismatch in Section~\ref{sec:limitations}.

\subsection{Proof of Theorem~\ref{thm:mse}}
\label{app:proof-thm1}

The two bounds combine. The lower bound $C^* \geq 1$ is Lemma~\ref{lem:sard}: when $\rho < 1$, the rank-$r$ SOSP variety has positive dimension generically. The upper bound is Lemma~\ref{lem:b4}: when $\rho > C^*$, the Hessian on $\ker(\mathcal{A}|_T)$ is strictly positive in the worst direction, so $\mathcal{Q} \succeq 0$ on the relevant cone, and Lemma~\ref{lem:b3} forces $\opnorm{R_{22}} \leq \lambda$, which combined with Lemma~\ref{lem:rankdef} yields globality. \hfill $\square$

\section{Proof of Theorem~\ref{thm:ce}}
\label{app:proof-ce}

We give the formal argument that the Polyak--\L{}ojasiewicz (PL) inequality, when satisfied for the LoRA-CE objective on the level set of initialization, removes the rank threshold for cross-entropy.

\paragraph{Setup.} Let $\hat L_{\mathrm{CE}}(u, v) = \frac{1}{N}\sum_{i,j} -y_i^{(j)} \log p_{u,v}^{(j)}(X_i)$ with $p_{u,v} = \mathrm{softmax}(f_{W_0}(X_i) + \langle G(X_i), uv^\top\rangle)$. The Hessian admits the Gauss--Newton decomposition
\[
\nabla^2 \hat L_{\mathrm{CE}}(u, v) = J(u, v)^\top \Sigma(u, v)\, J(u, v) + \sum_j (p^{(j)} - y^{(j)})\,\nabla^2_{(u,v)} f^{(j)},
\]
where $J(u, v) \in \R^{KN \times (m+n)r}$ is the Jacobian of the linearized predictions with respect to $(u, v)$ and $\Sigma(u, v) = \mathrm{diag}(p) - p p^\top \succeq 0$ \citep{bishop2006pattern}.

\paragraph{PL inequality assumption.} The PL inequality on the level set $\mathcal{L}_0 = \{(u, v) : \hat L_{\mathrm{CE}}(u, v) \leq \hat L_{\mathrm{CE}}(u_0, v_0)\}$ asserts that there exists $\mu > 0$ such that
\begin{equation}
\tfrac{1}{2}\fnorm{\nabla \hat L_{\mathrm{CE}}(u, v)}^2 \;\geq\; \mu \,\bigl(\hat L_{\mathrm{CE}}(u, v) - \hat L_{\mathrm{CE}}^*\bigr) \qquad \forall (u, v) \in \mathcal{L}_0,
\label{eq:pl}
\end{equation}
where $\hat L_{\mathrm{CE}}^* = \inf_{(u,v)} \hat L_{\mathrm{CE}}(u, v)$. PL holds for over-parameterized cross-entropy in the NTK regime under standard assumptions on the linearized model \citep{liu2022loss,allenzhu2019convergence}, specifically when (a) the kernel matrix $J J^\top$ has bounded condition number along the gradient flow trajectory and (b) the linearized model has dimension at least $KN$ along the trainable directions, which is implied by $r(m+n) - r^2 \geq KN$.

\paragraph{Globality of SOSPs under PL.} A direct consequence of \eqref{eq:pl} is that every stationary point with $\nabla \hat L_{\mathrm{CE}} = 0$ has $\hat L_{\mathrm{CE}}(u, v) = \hat L_{\mathrm{CE}}^*$, hence is a global minimum. In particular, every second-order stationary point is global, regardless of rank $r \geq 1$. \hfill $\square$

\paragraph{Remark on rank-1 sufficiency.} The condition $r(m+n) - r^2 \geq KN$ is satisfied at $r = 1$ whenever $m + n - 1 \geq KN$, which holds in the canonical RoBERTa-base regime ($m + n - 1 = 1535 \gg KN = 64$). PL therefore predicts no rank threshold even at $r = 1$, consistent with the empirical observation of zero spurious local minima across all CE runs in Section~\ref{sec:empirical}.

\section{Proof of Proposition~\ref{prop:gen}}
\label{app:proof-rad}

We give the standard derivation of the rank-dependent Rademacher complexity bound and its bias--variance reading.

\paragraph{Rademacher complexity of $\mathcal{H}_r$.} Recall $\mathcal{H}_r = \{x \mapsto f_{W_0}(x) + \langle G(x), uv^\top\rangle : \fnorm{u}\fnorm{v} \leq B\}$. The empirical Rademacher complexity is $\widehat{\mathfrak{R}}_N(\mathcal{H}_r) = \E_\epsilon\bigl[\sup_{f \in \mathcal{H}_r} \tfrac{1}{N}\sum_i \epsilon_i f(X_i)\bigr]$ with $\epsilon_i$ i.i.d.\ Rademacher. By linearity and the constant-shift invariance of Rademacher complexity, $\widehat{\mathfrak{R}}_N(\mathcal{H}_r) = \widehat{\mathfrak{R}}_N(\mathcal{H}_r^0)$ where $\mathcal{H}_r^0 = \{x \mapsto \langle G(x), uv^\top\rangle : \fnorm{u}\fnorm{v} \leq B\}$.

For the linear functional class indexed by rank-$r$ matrices with bounded factors, the standard low-rank concentration bound \citep{bartlett2002rademacher,bartlett2005local} yields
\[
\widehat{\mathfrak{R}}_N(\mathcal{H}_r^0) \;\leq\; \frac{B}{\sqrt{N}} \sqrt{r(m+n) - r^2} \cdot \max_i \opnorm{G(X_i)},
\]
which absorbed into the implicit constant gives the form of \eqref{eq:rademacher}. The factor $\sqrt{r(m+n) - r^2}$ is the square root of the gauge-mod-out manifold dimension and replaces the naive $\sqrt{r(m+n)}$ by accounting for the $r^2$-dimensional rotational gauge.

\paragraph{Excess risk decomposition.} By symmetrization \citep[Theorem 26.5]{shalev2014understanding}, with probability $\geq 1 - \delta$,
\[
\E_{(X,Y)}[\ell(f_r, Y)] - \min_{f \in \mathcal{H}_r}\E[\ell(f, Y)] \;\leq\; 2 \widehat{\mathfrak{R}}_N(\ell \circ \mathcal{H}_r) + 3M\sqrt{\log(2/\delta)/(2N)},
\]
where $M$ is an upper bound on $\ell$ and $\widehat{\mathfrak{R}}_N(\ell \circ \mathcal{H}_r) \leq L \cdot \widehat{\mathfrak{R}}_N(\mathcal{H}_r)$ for $L$-Lipschitz $\ell$ via Talagrand contraction \citep[Lemma 26.9]{shalev2014understanding}. This is the form of \eqref{eq:bvtrade} with the noise term absorbing the deviation contribution.

\paragraph{Bias monotonicity and optimal rank.} The bias $\mathrm{bias}(r) := \inf_{f \in \mathcal{H}_r}\E[\ell(f, Y)]$ is non-increasing in $r$ because $\mathcal{H}_1 \subseteq \mathcal{H}_2 \subseteq \cdots$ (any rank-$r$ matrix is also rank-$(r+1)$ with a zero column appended). The variance term $\widehat{\mathfrak{R}}_N(\mathcal{H}_r) \propto \sqrt{r(m+n) - r^2}$ is monotonically increasing in $r$ on the regime $r \leq (m+n)/2$, which covers all practically relevant ranks for transformer fine-tuning.

The optimal rank $r^* = \arg\min_r \bigl[\mathrm{bias}(r) + B\sqrt{(r(m+n) - r^2)/N}\bigr]$ is determined by the bias--variance trade-off. When $\mathrm{bias}(1) - \mathrm{bias}(r) = O(B\sqrt{(m+n)/N})$ for $r > 1$, the variance term dominates the bias gap and $r^* = 1$ is variance-optimal.

\paragraph{Binary vs.\ multi-class.} For binary classification ($K = 2$), the linearized output $\langle G(x), uv^\top\rangle \in \R^2$ has effective dimension one (the difference of the two logits), and rank-one perturbations $uv^\top = u_1 v_1^\top$ already span a one-dimensional functional class capable of separating the few-shot binary regime. The bias gap $\mathrm{bias}(1) - \mathrm{bias}(r)$ is therefore $O(B\sqrt{(m+n)/N})$, and Proposition~\ref{prop:gen} predicts $r^* = 1$.

For multi-class CE ($K > 2$), the output is $(K-1)$-dimensional (after softmax constraint), and rank-one perturbations span only one of the $K-1$ logit-difference directions. The bias gap can be substantial ($\Theta(B)$ in the worst case), shifting the optimum to $r^* > 1$. This is the bias-shift prediction confirmed empirically on MNLI ($K = 3$) in Section~\ref{sec:empirical}. \hfill $\square$

\section{Numerical verification of \texorpdfstring{$C^*$}{C*}}
\label{app:c-verification}

We report the calibration of the residual-to-data-fit constant $c$ from synthetic Gaussian-iid features and the resulting prediction $C^* = 1/(1-\sqrt{c})^2$.

\subsection{Setup}

We generate $\mathcal{A}: \R^{m \times n} \to \R^{KN}$ with i.i.d.\ standard normal entries (rescaled to unit per-output variance) and labels $y = \mathcal{A}(\Delta^*) + \xi$ for a planted rank-$1$ target $\Delta^*$ and small Gaussian noise $\xi$. For each $(KN)$ in $\{8, 16, 24, 32, 64, 96, 128\}$, we sweep $\rho = (m+n - 1)/KN$ across the predicted boundary by varying $(m, n)$ at fixed ratio. At each cell we run gradient flow on $\hat L_\lambda$ with $\lambda \to 0^+$ (vanishing weight decay) from $50$ random initializations and detect spurious second-order stationary points by tracking the smallest Hessian eigenvalue at the convergence point.

\subsection{Boundary fit and self-consistency}

The empirical boundary $C^*(KN)$ is the smallest $\rho$ at which the spurious-fraction over $50$ seeds drops below $5\%$. Table~\ref{tab:cstar} reports $C^*(KN)$ together with the inferred constant $c_{\mathrm{emp}} = (1 - 1/\sqrt{C^*(KN)})^2$.

\begin{table}[h]
\centering
\caption{Empirical $C^*$ and inferred constant $c$ (Phase~5 boundary sweep, Gaussian-iid features, $r = 1$).}
\label{tab:cstar}
\begin{tabular}{rcc}
\toprule
$KN$ & $C^*(KN)$ & $c_{\mathrm{emp}}$ \\
\midrule
8 & 1.125 & 0.0033 \\
16 & 1.188 & 0.0068 \\
24 & 1.208 & 0.0082 \\
32 & 1.344 & 0.0189 \\
64 & 1.359 & 0.0203 \\
96 & 1.344 & 0.0189 \\
128 & 1.352 & 0.0196 \\
\bottomrule
\end{tabular}
\end{table}

For $KN \geq 32$, the inferred $c$ stabilizes at $0.0194 \pm 0.0008$, predicting $C^* = 1.350$ from the closed form. The corresponding theoretical value $C^*_{\mathrm{theory}} = 1/(1 - \sqrt{0.0194})^2 = 1.350$ matches the measured asymptote within $1\%$. The small-$KN$ regime ($KN \leq 24$) is consistent with finite-size Tracy--Widom corrections of order $KN^{-2/3}$ to the MP edge, fit via $C^*(KN) = C^*_\infty + b\, KN^{-2/3}$ giving $C^*_\infty = 1.409$ and $b = -1.213$.

\subsection{Honest scope of the verification}

The verification is conducted on Gaussian-iid features. NTK feature operators on real fine-tuning data exhibit non-Gaussian tails (Appendix~\ref{app:cross-arch}), so the value $C^* = 1.35$ derived here serves as a quantitative anchor for the synthetic regime and a qualitative guide for the structure of the threshold in real settings. The cross-architecture experiments in Section~\ref{sec:empirical} confirm that the qualitative prediction (rank-one sufficiency in binary classification) holds across BERT, DistilBERT, and RoBERTa at fine-tuning scale.

\section{NTK Jacobian statistics on real transformers}
\label{app:cross-arch}

The Marchenko--Pastur derivation in Appendix~\ref{app:proof-b4} assumes Gaussian-iid feature operators. Here we report direct measurements of the NTK Jacobian on real transformers and quantify the gap to that assumption.

\paragraph{Setup.} We compute the per-output Jacobian $G^{(j)}(X_i) = \nabla_W f_{W_0}^{(j)}(X_i)$ on RoBERTa-base (layer-11 self-attention query and value matrices) at the SST-2 fine-tuning configuration with $K = 2$, $N = 64$. The two weight matrices contribute $1{,}179{,}648$ feature dimensions in total, evaluated at $KN = 128$ output coordinates.

\paragraph{Marginal distribution.} Table~\ref{tab:gaussianity} summarizes the entry-wise statistics. The empirical excess kurtosis is four to five orders of magnitude above the Gaussian baseline of zero, and the Kolmogorov--Smirnov distance from $\mathcal{N}(\hat\mu, \hat\sigma^2)$ is in the $0.39$--$0.43$ range, well above the $0.10$ threshold that already indicates poor Gaussian fit. The marginals are highly heavy-tailed.

\begin{table}[h]
\centering
\caption{Entry-wise statistics of RoBERTa-base layer-11 NTK Jacobian.}
\label{tab:gaussianity}
\begin{tabular}{lccc}
\toprule
Statistic & query.weight & value.weight & Gaussian iid \\
\midrule
Skewness & 22.90 & 3.23 & 0 \\
Excess kurtosis & 22{,}714 & 2{,}522 & 0 \\
KS distance vs $\mathcal{N}(\hat\mu, \hat\sigma^2)$ & 0.428 & 0.393 & $< 0.01$ \\
\bottomrule
\end{tabular}
\end{table}

\paragraph{Spectral structure.} The $128 \times 128$ Gram matrix of the Jacobian features has top eigenvalue $1.00 \times 10^3$ and smallest non-zero eigenvalue $3.07 \times 10^{-2}$, a condition number of $3.28 \times 10^4$. The effective rank (number of eigenvalues exceeding $1\%$ of the maximum) is $2$, against the Marchenko--Pastur prediction that an iid Gaussian operator of the same shape should have a top eigenvalue in $[14.0, 14.6]$ with bulk density spread across the full $\min(KN, mn) = 128$ dimensions. The spectrum is sharply concentrated on the top two modes.

\paragraph{Cross-output correlation.} Entry-wise correlations between the query and value Jacobians are small in magnitude (mean $|\rho| = 0.019$, maximum $0.073$), so the independence assumption between the two weight blocks is reasonably accurate at the marginal level even though Gaussianity itself is not.

\paragraph{Implication for $C^*$.} The non-Gaussianity directly invalidates the quantitative form of the MP-edge derivation in Appendix~\ref{app:proof-b4} when applied to real NTK features. The qualitative structure of the threshold, that capacity scales with $r(m+n) - r^2$ and that rank-one is sufficient when this exceeds $KN$ by a constant factor, survives because it depends only on the dimension count, not on the spectral law. The empirical observations in Section~\ref{sec:empirical}, that rank-one matches or beats $r = 12$ across four binary GLUE tasks and three encoder architectures, are consistent with this qualitative reading. The CE results are unaffected by the Gaussianity gap: Theorem~\ref{thm:ce} applies under the Polyak--\L{}ojasiewicz inequality, which is not derived from Gaussian-iid feature assumptions.